\pdfoutput=1

\documentclass[11pt]{article}

\usepackage{acl}

\usepackage{times}
\usepackage{latexsym}
\usepackage{amssymb}
\usepackage[T1]{fontenc}
\usepackage[ruled,linesnumbered]{algorithm2e}

\usepackage[utf8]{inputenc}
\usepackage{amsmath} 

\usepackage{microtype}
\usepackage{spverbatim}

\usepackage{booktabs}
\usepackage{graphicx}
\usepackage[normalem]{ulem}
\useunder{\uline}{\ul}{}

\DeclareMathOperator*{\argmax}{argmax}

\title{Inference to the Best Explanation in Large Language Models}

\author{Dhairya Dalal$^{1}$ , Marco Valentino$^2$, Andr\'e Freitas$^{2,3,4}$, Paul Buitelaar$^{1,5}$ \\
$^{1}$ SFI Centre for Research and Training in Artificial Intelligence, University of Galway, Ireland \\
$^{2}$ Idiap Research Institute, Switzerland\\
$^{3}$ Department of Computer Science, University of Manchester, UK\\
$^{4}$ National Biomarker Centre, CRUK-MI, University of Manchester, UK\\
$^{5}$ Insight SFI Research Centre for Data Analytics, University of Galway, Ireland  \\
Email: d.dalal1@universityofgalway.ie
}

\begin{document}
\maketitle
\begin{abstract}

While Large Language Models (LLMs) have found success in real-world applications, their underlying explanatory process is still poorly understood. This paper proposes \textit{IBE-Eval}, a framework inspired by philosophical accounts on \emph{Inference to the Best Explanation (IBE)} to advance the interpretation and evaluation of LLM explanations. \textit{IBE-Eval} estimates the plausibility of natural language explanations through a combination of explicit logical and linguistic features including: \emph{consistency}, \emph{parsimony}, \emph{coherence}, and \emph{uncertainty}. Extensive experiments are conducted on \emph{Causal Question Answering (CQA)}, where \textit{IBE-Eval} is tasked to select the most plausible causal explanation amongst competing ones generated by the LLM (e.g. GPT 3.5 or LLaMA 2). The experiments reveal that \textit{IBE-Eval} can successfully identify the best explanation with up to 77\% accuracy ($\approx 27\%$ above random), improving upon a GPT 3.5-as-a-judge baseline ($\approx+17\%$) while being intrinsically more efficient and interpretable. Additional analysis suggests that, despite LLM-specific variances, generated explanations tend to conform to IBE criteria and that \textit{IBE-Eval} is significantly correlated with human judgment, opening up opportunities for future development of automated explanation verification tools.

\end{abstract}

\begin{figure*}[ht!]
    \centering
    \includegraphics[width=.95\textwidth]{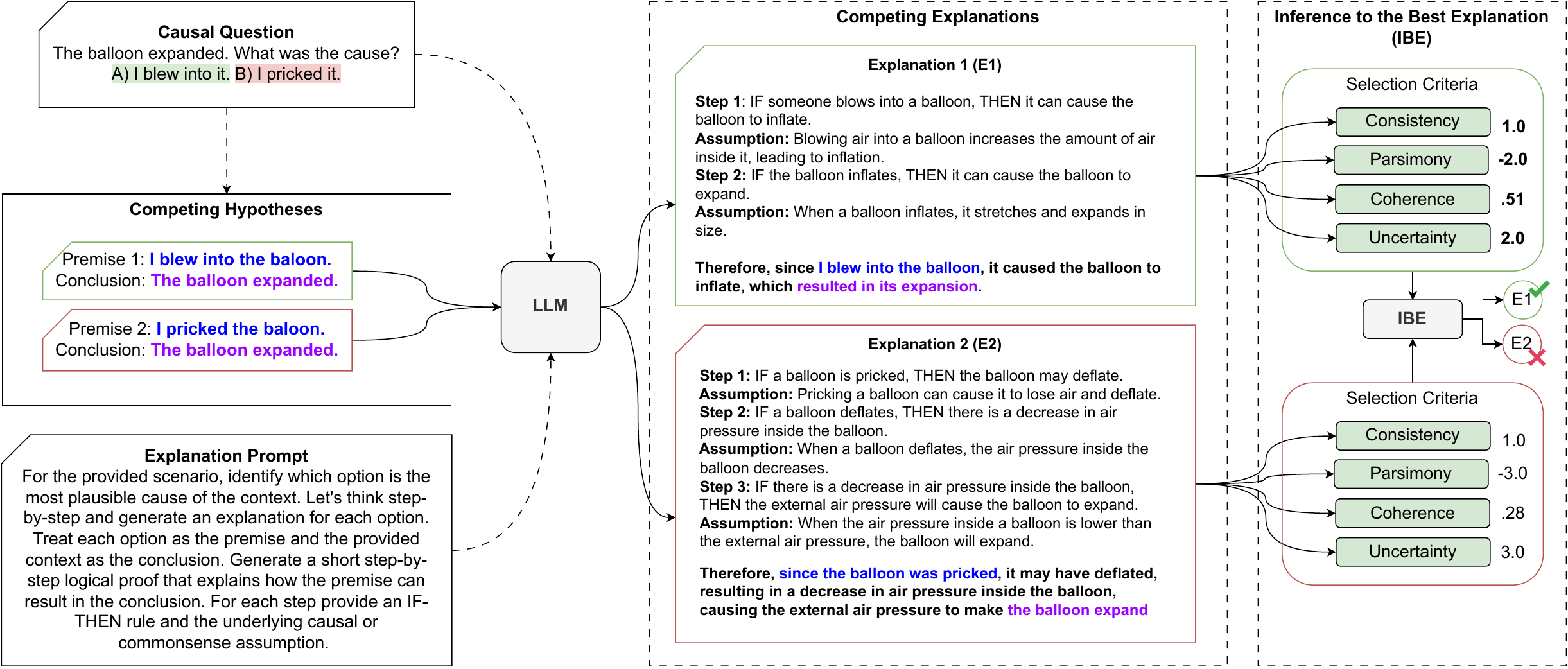}
    \caption{\textit{IBE-Eval} qualifies LLM-generated explanations with a set of logical and linguistic selection criteria to identify the most plausible hypothesis. The corresponding explanation for each hypothesis is evaluated across the IBE criteria of logical consistency, parsimony, internal coherence, and linguistic uncertainty. A final plausibility score is computed across those features and the hypothesis with highest score is identified as the best explanation. }
    \label{fig:overall_framework}
\end{figure*}

\section{Introduction}

Large Language Models (LLMs) such as OpenAI's GPT \cite{brown2020language} and LLaMA \cite{touvron2023llama} have been highly effective across a diverse range of language understanding and reasoning tasks \cite{liang2023holistic}. While LLM performances have been thoroughly investigated across various benchmarks \cite{superglue, big-bench, eval-harness, touvron2023llama}, the principles and properties behind their step-wise reasoning process are still poorly understood \cite{valentino-etal-2021-natural}. LLMs are notoriously black-box and can be difficult to interpret \cite{dl-interp-survey, danilevsky-etal-2020-survey}. Moreover, the commercialization of LLMs has led to strategic secrecy around model architectures and training details \cite{vice, wired}. Finally, LLMs are susceptible to hallucinations and adversarial perturbations \cite{geirhos2020shortcut,camburu2020make}, often producing plausible but factually incorrect answers \cite{survey-hallucinations, huang2023survey}. As the size and complexity of LLM architectures increase, the systematic study of generated explanations becomes crucial to better interpret and validate the LLM's internal inference and reasoning processes \cite{wei2022chain,lampinen2022can,huang2022towards}. 

The automatic evaluation of natural language explanations presents several challenges \cite{atanasova-etal-2023-faithfulness,camburu2020make}. Without resource-intensive annotation \cite{wiegreffe2021teach,thayaparan2020survey,dalvi2021explaining,camburu2018snli}, explanation quality methods tend to rely on either weak supervision, where the identification of the correct answer is taken as evidence of explanation quality, or require the injection of domain-specific knowledge \cite{quan2024enhancing}. In this paper, we seek to better understand the LLM explanatory process through the investigation of explicit linguistic and logical properties. While explanations are hard to formalize due to their open-ended nature, we hypothesize that they can be analyzed as linguistic objects, with measurable features that can serve to define criteria for assessing their quality. 

Specifically, this paper investigates the following overarching research question: \emph{``Can the linguistic and logical properties associated with LLM-generated explanations be used to qualify the models' reasoning process?''}. To this end, we propose an interpretable framework inspired by philosophical accounts of abductive inference, also known as \emph{Inference to the Best Explanation (IBE)} - i.e. the process of selecting among competing explanatory theories \cite{lipton2017inference}. In particular, we aim to measure the extent to which LLM-generated explanations conform to IBE expectations when attempting to identify the most plausible explanation. To this end, we present \textit{IBE-Eval}, a framework designed to estimate the plausibility of natural language explanations through a set of explicit logical and linguistic features, namely: \emph{logical consistency}, \emph{parsimony}, \emph{coherence}, and \emph{linguistic uncertainty}. 

To evaluate the efficacy of \textit{IBE-Eval}, we conduct extensive experiments in the multiple-choice Causal Question Answering (CQA) setting. The overall results and contributions of the paper can be summarized as follows: 





\begin{enumerate}
    \item To the best of our knowledge, we are the first to propose an interpretable framework inspired by philosophical accounts on Inference to the Best Explanation (IBE) to automatically assess the quality of natural language explanations.
    \item We propose \textit{IBE-Eval}, a framework that can be instantiated with external tools for the automatic evaluation of LLM-generated explanations and the identification of the best explanation in a multiple-choice CQA setting.
    \item We provide empirical evidence that LLM-generated explanations tend to conform to IBE expectations with varying levels of statistical significance correlated to the LLM's size. 
    \item We additionally find that uncertainty, parsimony, and coherence are the best predictors of plausibility and explanation quality across all LLMs. However, we also find that the LLMs tend to be strong rationalizers and can produce logically consistent explanations even for less plausible candidates, making the consistency metric less effective in practice.   
    \item \textit{IBE-Eval} can successfully identify the best explanation supporting the correct answers with up to 77\% accuracy (+$\approx 27\%$ above random and +$\approx17\%$ over GPT 3.5-as-a-Judge baselines)
    \item \textit{IBE-Eval} is significantly correlated with human judgment, outperforming a GPT3.5-as-a-Judge baseline in terms of alignment with human preferences. 
\end{enumerate}

For reproducibility, our code is made available on Github\footnote{\url{https://github.com/dhairyadalal/IBE-eval}} to encourage future research in the field.

\section{Inference to the Best Explanation (IBE)}

Explanatory reasoning is a distinctive feature of human rationality underpinning problem-solving and knowledge creation in both science and everyday scenarios \cite{lombrozo2012explanation,deutsch2011beginning}. Accepted epistemological accounts characterize the creation of an explanation as composed of two distinct phases: conjecturing and criticism \cite{popper2014conjectures}. The explanatory process always involves a conflict between plausible explanations, which is typically resolved through the criticism phase via a selection process, where competing explanations are assessed according to a set of criteria 
such as parsimony, coherence, unification power, and  
hardness to variation \cite{lipton2017inference,harman1965inference,mackonis2013inference,thagard1978best,thagard1989explanatory,kitcher1989explanatory,valentino2022scientific}.

As LLMs become interfaces for natural language explanations, epistemological frameworks offer an opportunity for developing criticism mechanisms to understand the explanatory process underlying state-of-the-art models. To this end, this paper considers an LLM as a conjecture device producing linguistic objects that can be subject to criticism. In particular, we focus on a subset of criteria that can be computed on explicit linguistic and logical features, namely: \textit{consistency}, \textit{parsimony}, \textit{coherence}, and \textit{uncertainty}.

To assess the LLM's alignment to such criteria, we focus on the task of selecting among competing explanations in a multiple-choice CQA setting (Figure \ref{fig:overall_framework}). Specifically, given a set of competing hypotheses (i.e. the multiple-choice options), $H = \{h_1, h_2, \ldots, h_n\}$, we prompt the LLM to generate plausible explanations supporting each hypothesis (Section \ref{sec:prompting}). Subsequently, we adopt the proposed IBE selection criteria to assess the quality of the generated explanations (Section \ref{sec:explanation_criteria}). \textit{IBE-Eval} computes an explanation plausibility score derived from the linear combination of the computed selection criteria. The explanation with the highest score is selected as the predicted answer and additionally assessed as to the extent to which the observable IBE features are correlated with QA accuracy. We hypothesize that {IBE-Eval} will produce higher scores for the explanation associated with the correct answer and that the IBE criteria should meaningfully differentiate between competing explanations.                                           

\section{Explanation Generation}
\label{sec:prompting}
For the first stage, the LLM is prompted to generate competing explanations for the hypotheses using a modified Chain-of-Thought (CoT) prompt \cite{cot-prompting}. Specifically, the COT prompt is modified to instruct the LLM to produce an explanation for each competing hypothesis (see Figure \ref{fig:overall_framework}). We adopt a methodology similar to \citet{valentino-etal-2021-natural}, where the generated explanation is constrained into an entailment form for the downstream IBE evaluation. In particular, we posit that a valid explanation should demonstrate an entailment relationship between the premise and conclusion which are derived from the question-answer pair. 


To elicit logical connections between explanation steps and facilitate subsequent analysis, the LLM is constrained to use weak syllogisms expressed as If-Then statements. Additionally, the LLM is instructed to produce the associated causal or commonsense assumption underlying each explanation step. This output is then post-processed to extract the explanation steps and supporting knowledge for evaluation via the IBE selection criteria. Additional details and examples of prompts are reported in Appendix \ref{sec:appendix-prompting}. 



\section{Linguistic \& Inference Criteria}
To perform IBE, we investigate a set of criteria that can be automatically computed on explicit logical and linguistic features, namely: \textit{consistency}, \textit{parsimony}, \textit{coherence}, and \textit{uncertainty}.

\label{sec:explanation_criteria}

\paragraph{Consistency.}
Consistency aims to verify whether the explanation is logically valid. Given a hypothesis, comprised of a premise $p_i$, a conclusion $c_i$, and an explanation consisting of a set of If-Then statements $E = {s_1, ..., s_i}$, we define $E$ to be logically consistent if $p_i \cup E \vDash c_i$. Specifically, an explanation is logically consistent if it is possible to build a deductive proof linking premise and conclusion.

To evaluate logical consistency, we leverage external symbolic solvers along with autoformalization - i.e., the translation of natural language into a formal language \cite{wu2022autoformalization}. Specifically, the hypotheses and explanations are formalized into a Prolog program which will attempt to generate a deductive proof via backward chaining \cite{weber2019nlprolog}. 

To perform autoformalization, we leverage the translation capabilities of GPT 3.5. Specifically, we instruct GPT 3.5 to convert each IF-Then explanation step from the generated explanation into an implication rule and the premise statement into grounding atoms. On the other end, the entailment condition and the conclusion are used to create a Prolog query. The query instructs the Prolog solver to attempt to find a path through the implications rules such that the conclusion be directly connected to the premise. Further details about the autoformalization process can be found in Appendix \ref{sec:appendix-autoformalization}.

After autoformalization, following recent work on neuro-symbolic integration for LLM explanations \cite{quan2024enhancing}, we adopt an external Prolog solver for entailment verification\footnote{\url{https://github.com/neuro-symbolic-ai/explanation_based_ethical_reasoning}}. The explanation is considered consistent if the Prolog solver can satisfy the query and successfully build a deductive proof. Technical details can be found in Appendix \ref{sec:appendix-consistency}.

\paragraph{Parsimony.}
The parsimony principle, also known as Ockham's razor, favors the selection of the simplest explanation consisting of the fewest elements and assumptions \cite{sober-parsimony}. Epistemological accounts posit that an explanation with fewer assumptions tends to leave fewer statements unexplained, improving specificity and alleviating the infinite regress \cite{thagard1978best}. Further, parsimony is an essential feature of causal interpretability, as only parsimonious solutions are guaranteed to reflect causation in comparative analysis \cite{baumgartner2015parsimony}. 
In this paper, we adopt two metrics as a proxy of parsimony, namely \emph{proof depth}, and \emph{concept drift}. 
Proof depth, denoted as \( Depth \), is defined as the cardinality of the set of rules, \( R \), required by the Prolog solver to connect the conclusion to the premise via backward chaining. 
Let \( h \) be a hypothesis candidate composed of a premise \( p \) and a conclusion \( c \), and let \( E\) be a formalized explanation represented as a set of rules \( R'\). The proof depth is the number of rules $|R|$, with $R \subseteq R'$, traversed during backward chaining to connect the conclusion \( c\) to the premise \(p\):
\[ Depth(h) = |R| \]

Concept drift, denoted as \( Drift \), is defined as the number of additional concepts and entities, outside the ones appearing in the hypothesis (i.e., premise and conclusion), that are introduced by the LLM to support the entailment. For simplicity, we consider nouns as concepts. Let $N = \{Noun_{p}, Noun_{c}, Noun_{E} \}$ be the unique nouns found in the premise, conclusion, and explanation steps. Concept drift is the cardinality of the set difference between the nouns found in the explanation and the nouns in the hypothesis:
\[ Drift(h) = |Noun_{E} - (Noun_{p} \cup Noun_{c})| \]
Intuitively, the parsimony principle would predict the most plausible hypothesis as the one supported by an explanation with the smallest observed proof depth and concept drift. Implementation details can be found in Appendix \ref{sec: appendix-parsimony}.

\paragraph{Coherence.}
\textit{Coherence} attempts to measure the logical validity at the level of the specific explanation steps. An explanation can be formally consistent on the surface while still including implausible or ungrounded intermediate assumptions. Coherence evaluates the quality of each intermediate If-Then implication by measuring the entailment strength between the If and Then clauses. To this end, we employ a fine-tuned natural language inference (NLI) model. Formally, let \(S\) be a set of explanation steps, where each step \(s\) consists of an If-Then statement, \(s = (If_s, Then_s)\). For a given step \(s_i\), let \(ES(s_i)\) denote the entailment score obtained via the NLI model between $If_s$ and $Then_s$ clauses. The step-wise entailment score $SWE(S)$ is then calculated as the averaged sum of the entailment scores across all explanation steps $|S|$:
\[
\text{SWE}(S) = \frac{1}{|S|}\sum_{i=1}^{|S|} \text{ES}(s_i)
\]
We hypothesize that the LLM should generate a higher coherence score for more plausible hypotheses, as such explanations should exhibit stronger step-wise entailment. Additional details can be found in Appendix \ref{sec: appendix-coherence }.

\paragraph{Uncertainty.}
Finally, we consider the linguistic certainty expressed in the generated explanation as a proxy for plausibility. Hedging words such as \textit{probably}, \textit{might be}, \textit{could be}, etc typically signal ambiguity and are often used when the truth condition of a statement is unknown or improbable. \citet{pei-jurgens-2021-measuring} found that the strength of scientific claims in research papers is strongly correlated with the use of direct language. In contrast, they found that the use of hedging language suggested that the veracity of the claim was weaker or highly contextualized.

To measure the linguistic uncertainty ($UC$) of an explanation, we consider the explanation's underlying assumptions ($A_i$) and the overall explanation summary ($S$). The linguistic uncertainty score is extracted using the fine-tuned sentence-level RoBERTa model from \citet{pei-jurgens-2021-measuring}. The overall linguistic uncertainty score ($UC_{\text{overall}}$) is the sum of the assumption and explanation summary scores:

\[
UC_{\text{overall}} = UC(A) + UC(S)
\]
Where $UC(A)$ is the sum of the linguistic uncertainty scores ($UC(A)$) across all the assumptions $|A|$ associated with each explanation step $i$:

\[
UC(A) = \sum_{i=1}^{|A|} UC(a_i)
\]
and linguistic uncertainty of the explanation summary $UC(S)$. We hypothesize that the LLM will use more hedging language when explaining the weaker hypothesis resulting in a higher uncertainty score. Further details can be found in Appendix \ref{sec: appendix-uncertainty}.

\subsection{Inference to Best Explanation}
After the IBE criteria are computed for each competing hypothesis, they are used to generate the final explanation plausibility score. We define a simple linear regression model $\theta(\cdot)$, which was fitted on a small set of training examples consisting of extracted IBE features to predict the probability that an explanation $E_i$ corresponds to the correct answer. Specifically, we employ \textit{IBE-Eval} to score each generated explanation independently and then select the final answer $a$ via argmax:

\[
a = \argmax_i [\theta(E_1),\ldots, \theta(E_n)]
\]

Additional details can be found in Appendix \ref{sec: appendix-ibe}.

\begin{figure*}[t]
    \centering
    \includegraphics[width=0.9\textwidth]{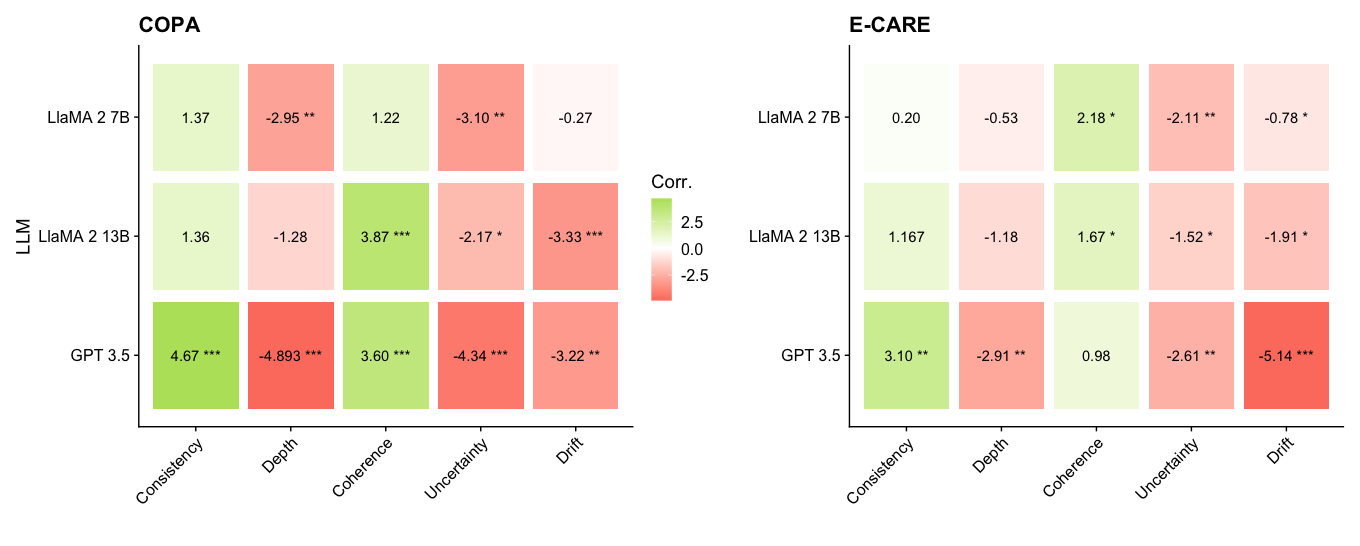}
    \caption{ A regression analysis measuring the correlation between IBE criteria and question accuracy. All the LLMs tend to conform to IBE expectations with GPT 3.5 exhibiting the most consistent and significant alignment. Linguistic uncertainty is the strongest IBE predictor for explanation quality, where higher uncertainty is negatively correlated with question accuracy. Statistical significance is noted as: ‘***’ p < 0.001, ‘**’ p < 0.01 ‘*’ p < 0.05.}
    \label{fig:correlation}
\end{figure*}

\begin{table*}[t!]
\centering
\small
\begin{tabular}{@{}lccc|ccc@{}}
\toprule
                         & \multicolumn{3}{c|}{\textbf{COPA}}                                    & \multicolumn{3}{c}{\textbf{E-CARE}}                                   \\ \midrule
                         & \textbf{GPT 3.5}      & \textbf{LlaMA 2 13B}  & \textbf{LlaMA 2 7B}   & \textbf{GPT 3.5}      & \textbf{LlaMA 2 13B}  & \textbf{LlaMA 2 7B}   \\ \midrule
{\ul \textbf{Baselines}}    &     &     &     &     &     &     \\
GPT3.5 Judge               & .59 & .47 & .63 & .43 & .61 & .52 \\
\textit{\textbf{Human}}  & \textit{\textbf{.95}} & \textit{\textbf{1.0}} & \textit{\textbf{.91}} & \textit{\textbf{.90}} & \textit{\textbf{.91}} & \textit{\textbf{.92}} \\ \midrule
{\ul \textbf{IBE Features}} &     &     &     &     &     &     \\
Consistency                 & .51 & .52 & .55 & .54 & .54 & .54 \\
Depth (Parsimony)           & .67 & .53 & .63 & .66 & .56 & .54 \\
Drift (Parsimony)           & .67 & .63 & .58 & .66 & .57 & .57 \\
Coherence                   & .66 & .66 & .56 & .56 & .57 & .59 \\
Linguistic Uncertainty      & .70 & .65 & .61 & .59 & .56 & .60 \\ \midrule
{\ul \textbf{Composed Model}}     &     &     &     &     &     &     \\
Random                      & .50 & .50 & .50 & .50 & .50 & .50 \\
+ Consistency               & .51 & .52 & .55 & .54 & .54 & .54 \\
+ Depth                     & .67 & .53 & .63 & .66 & .56 & .56 \\
+ Drift                     & .70 & .65 & .65 & .72 & .66 & .65 \\
+ Coherence                 & .73 & .71 & .69 & .73 & .68 & .69 \\
+ Linguistic Uncertainty & \textbf{.77}          & \textbf{.74}          & \textbf{.70}          & \textbf{.74}          & \textbf{.70}          & \textbf{.73}          \\ \bottomrule
\end{tabular}
\caption{An ablation study and evaluation of the IBE criteria and the composed \textit{IBE-Eval} model. \textit{IBE-Eval} outperforms the GPT 3.5 Judge baseline by an average of +17.5\% across all all models and tasks.}
\label{tab:ablation-table}
\end{table*}

\section{Experimental Setting}

Causal Question-Answering (CQA) requires reasoning about the causes and effects given an event description. We specifically consider the task of cause and effect prediction in a multiple-choice setting, where given a question and two candidate answers, the LLM must decide which is the most plausible cause or effect. Causal reasoning is a challenging task as the model must both possess commonsense knowledge about causal relationships and consider the event context which would make one option more plausible than the other. For our experiments, we use the Choice of Plausible Alternatives (COPA) \cite{gordon-etal-2012-semeval} and E-CARE \cite{du-etal-2022-e} datasets.

\paragraph{COPA.} 
COPA is a multiple-choice commonsense causal QA dataset consisting of 500 train and test examples that were manually generated. Each multiple-choice example consists of a question premise and a set of answer candidates which are potential causes or effects of the premise.
COPA is a well-established causal reasoning benchmark that is both a part of SuperGlue \cite{superglue} and the CALM-Bench \cite{dalal-etal-2023-calm}. 

\paragraph{E-CARE.}
E-CARE is a large-scale multiple-choice causal crowd-sourced QA dataset consisting of ~15K train and ~2k test examples. Similar to COPA, the task requires the selection of the most likely cause or effect provided an event description. We randomly sample 500 examples from the E-CARE test set for our experiments. 

\paragraph{LLMs.}
We consider GPT-Turbo-3.5, LLaMA 2 13B, and LLaMA 2 7B for all experiments. GPT 3.5 is a proprietary model \cite{brown2020language} and is highly effective across a wide range of natural language reasoning tasks \cite{laskar2023systematic}. We additionally evaluate the open-source LLaMA 2 model \cite{touvron2023llama}. We consider both the 13B and 7B variants of Llama 2 as both are seen as viable commodity GPT alternatives and have been widely adopted by the research community for LLM benchmarking and evaluation.   

\paragraph{Baselines.} We employ LLM-as-a-Judge \cite{Zheng2023JudgingLW} and human evaluators as baseline methods for the selection of the best explanation in the CQA setting. \citep{Zheng2023JudgingLW} found LLMs can align with human judgment and be utilized for automated evaluation and judgment. We specifically uses GPT 3.5 as the LLM judge. For each CQA example, we present the judges with two competing explanations generated by the target LLM. The judge is asked to identify the best and most plausible explanation. Additional details about the baselines can be found in Appendix \ref{sec:appendix-llm-judge}.

\section{Preliminary Analysis}
We conduct a preliminary analysis as a sanity check to measure the extent to which LLMs generate \emph{self-evident or tautological} explanations - i.e., explanations that simply restate the premises and conclusions. Tautological explanations present a risk for \textit{IBE-Eval } as the metrics would be theoretically uninformative if the LLM adopts the tested causal relation as the explanation itself (e.g. A → B) without providing additional supporting statements.

We consider the \emph{parsimony} metric to compute the percentage of explanations with \emph{proof depth} equal to 1 (i.e, explanations containing only one inference step) and \emph{concept drift} equal to 0 (i.e. no additional concepts other than the ones stated in premises and conclusions appear in the explanation). In such cases, the LLM is effectively generating a self-evident or tautological explanation.

\textbf{We found that about 2\% of the cases consist of self-evident explanations}. For GPT 3.5, LLaMA 2 13B, and LLaMA 2 7B, 2\% of the generated explanations exhibit a concept drift of 0, and on average 1.5\% of the explanations have a proof depth of 1. We then conducted an error analysis to evaluate the cases where \textit{IBE-Eval} selected a self-evident explanation as the best one. \textbf{Across all LLMs, less than 0.1\% of the errors were caused by the selection of such explanations}. Our analysis suggests that the impact of self-evident explanations is not significant and that the IBE framework can be robustly applied to identify such cases.

\section{Results}
To assess the LLM's alignment with the proposed IBE framework and evaluate the efficacy of \textit{IBE-Eval}, we run a regression analysis and conduct a set of ablation studies to evaluate the relationship between IBE and question accuracy. The main results are presented in Figure \ref{fig:correlation} and Table \ref{tab:ablation-table}. 

Our regression analysis finds that the IBE criteria are generally consistent across the LLMs as demonstrated by similar correlation patterns found on both the COPA and E-CARE tasks (Figure \ref{fig:correlation}). GPT 3.5 exhibits the strongest alignment with IBE expectations as we observe nearly all the IBE criteria have statistically significant and directionally aligned correlations across both tasks. \textbf{Thus our proposed IBE criteria can serve as promising build blocks for future work on automated explanation evaluation. }  

In Table \ref{tab:ablation-table} we evaluate the accuracy of the IBE criteria and \textit{IBE-Eval} in selecting the most plausible explanation in the CQA setting. We find that though independently the IBE criteria are generally limited in their ability to identify the more plausible explanation - they still outperform the GPT-3.5-as-a-judge baseline. \textbf{\textit{IBE-Eval}, which considers all IBE criteria, improves the ability to select the best explanation by 17\% over both the GPT 3.5-as-a-judge and random baselines.} We can achieve up to 77\% accuracy utilizing just the extracted IBE criteria demonstrating IBE's potential value for automatic explanation evaluation. 

Next, we explore each explanation feature in further detail to better understand the variances across the IBE criteria and LLMs. 

\begin{figure}[t!]
    \centering
    \includegraphics[width=.5\textwidth]{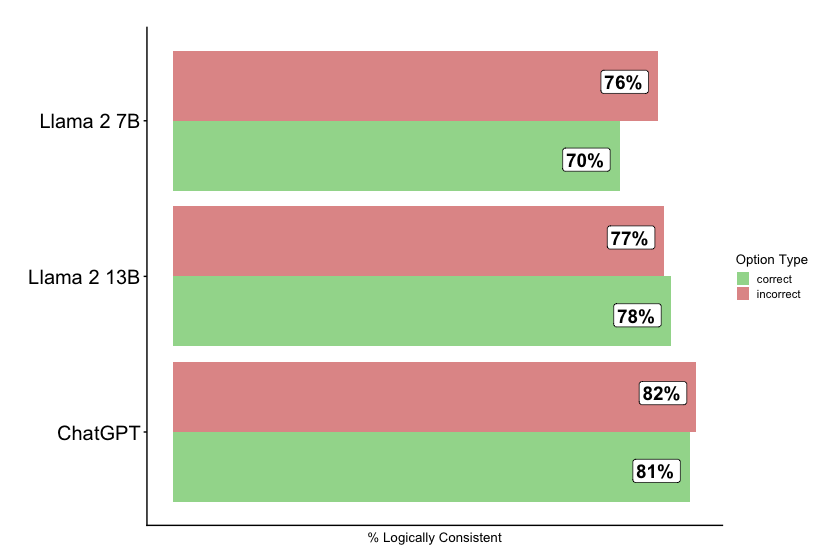}
    \caption{An evaluation of explanation consistency. LLMs are strong rationalizers and can generate logically consistent explanations at equal rates for explanations associated with both correct and incorrect answers options.}
    \label{fig:proof-consistency}
\end{figure}

\paragraph{Consistency. }  
We find that the LLMs are surprisingly strong conjecture models. The LLMs can generate logically consistent explanations for any hypothesis as observed by similar consistency scores for correct and incorrect (Figure \ref{fig:proof-consistency}) explanations. Moreover, we observe that consistency tends to be a statistically insignificant predictor for the LLaMA models. Therefore, we conclude that \textbf{evidence of logical consistency provides a limited signal for plausibility and is better understood in the context of other IBE criteria}. For the incorrect candidate explanations, we find that LLMs over-rationalize and introduce additional premises to demonstrate entailment in their explanations.

\paragraph{Parsimony.}
\begin{figure}[t]
    \centering
    \includegraphics[width=.5\textwidth]{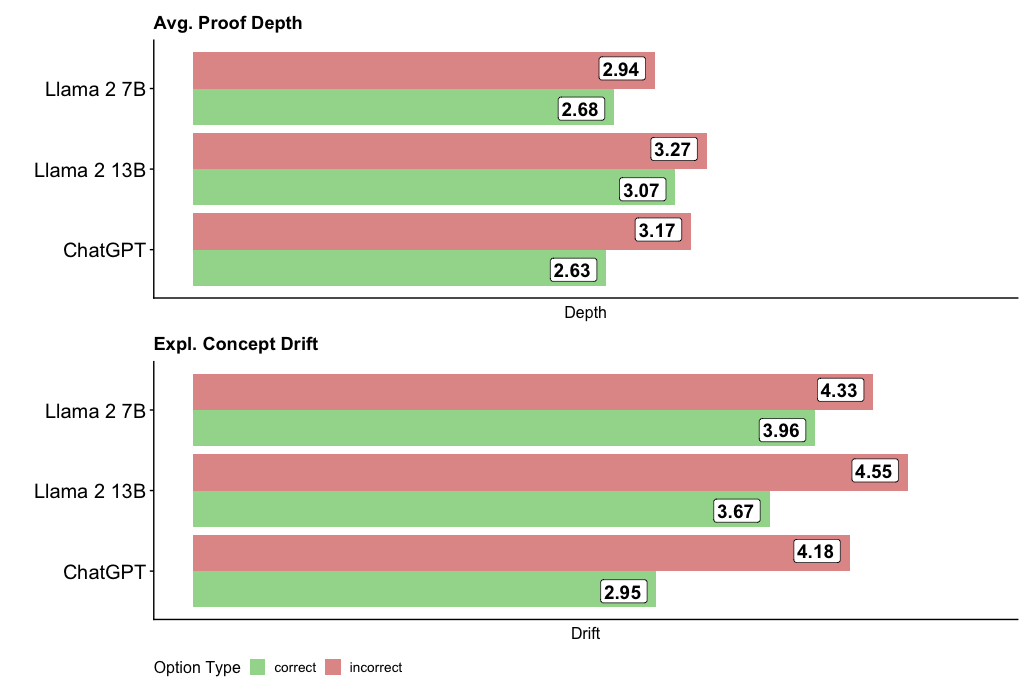}
    \caption{Explanation parsimony is evaluated using proof depth and concept drift. Both metrics are consistently lower for explanations supporting the correct answers suggesting that LLMs are able to generate efficient explanations for the more plausible hypothesis. }
    \label{fig:concept-drift}
\end{figure}

The results suggest that parsimony has a more consistent effect and is a better predictor of explanation quality. \textbf{We observe negative correlations between proof depth, concept drift, and question-answering accuracy, suggesting that LLMs tend to introduce more concepts and explanation steps when explaining less plausible hypotheses}. On average, we found the depth and drift to be 6\% and 10\% greater for the incorrect option across all LLMs (Figure \ref{fig:concept-drift}). Moreover, the results suggest that \textbf{as the LLM parameter size increases, the tendency to over-rationalize increases as well}. This is attested by the fact that the average difference in depth and drift is the greatest in GPT 3.5, suggesting that the model tends to find the most efficient explanations for stronger hypotheses and articulates explanations for weaker candidates. Finally, we found that the \textbf{LLaMA models tend to generate more complex explanations overall}, with LLaMA 2 13B exhibiting the largest concept drift for less plausible hypotheses. \textbf{The parsimony criterion supports the IBE predictive power with an average of 14\% improvement over consistency}. 

\paragraph{Coherence.}
\begin{figure}[t]
    \centering
    \includegraphics[width=.5\textwidth]{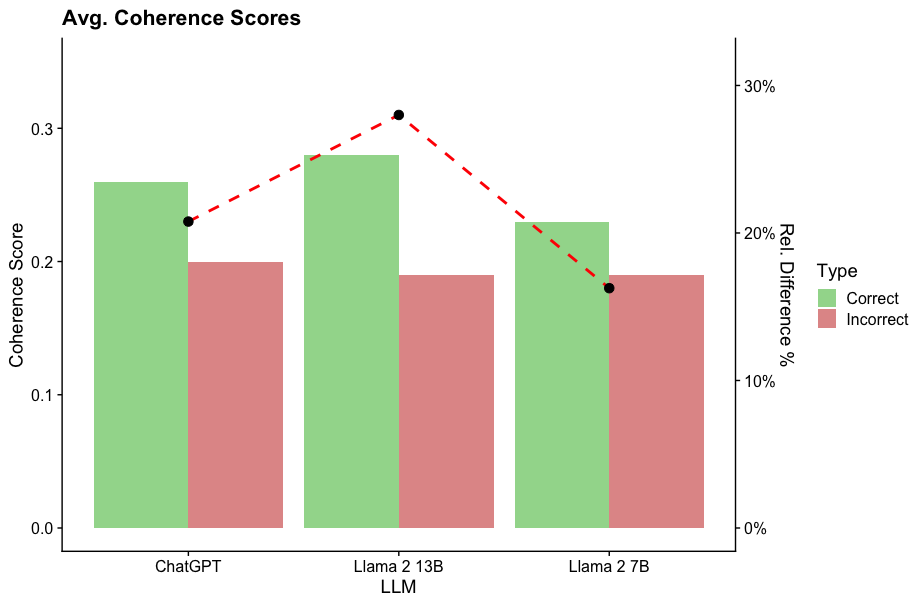}
    \caption{An evaluation of the explanation coherence and question accuracy.The average coherence score is consistently higher for explanations corresponding to the correct hypotheses across the LLMs.}
    \label{fig:coherence}
\end{figure}

Similarly to parsimony, \textbf{we found coherence to be a better indicator of explanation quality being statistically significant for both GPT 3.5 and Llama 2 13B on COPA and both Llama 2 models on E-Care}. We found that the average coherence score is consistently greater for the stronger hypothesis across all LLMs and datasets (see Figure \ref{fig:coherence}). \textbf{Both GPT and Llama 2 13B exhibit a higher relative difference between the correct and incorrect hypotheses in contrast to Llama 2 7B}. 

\paragraph{Uncertainty.}
\begin{figure}[t]
    \centering
    \includegraphics[width=.5\textwidth]{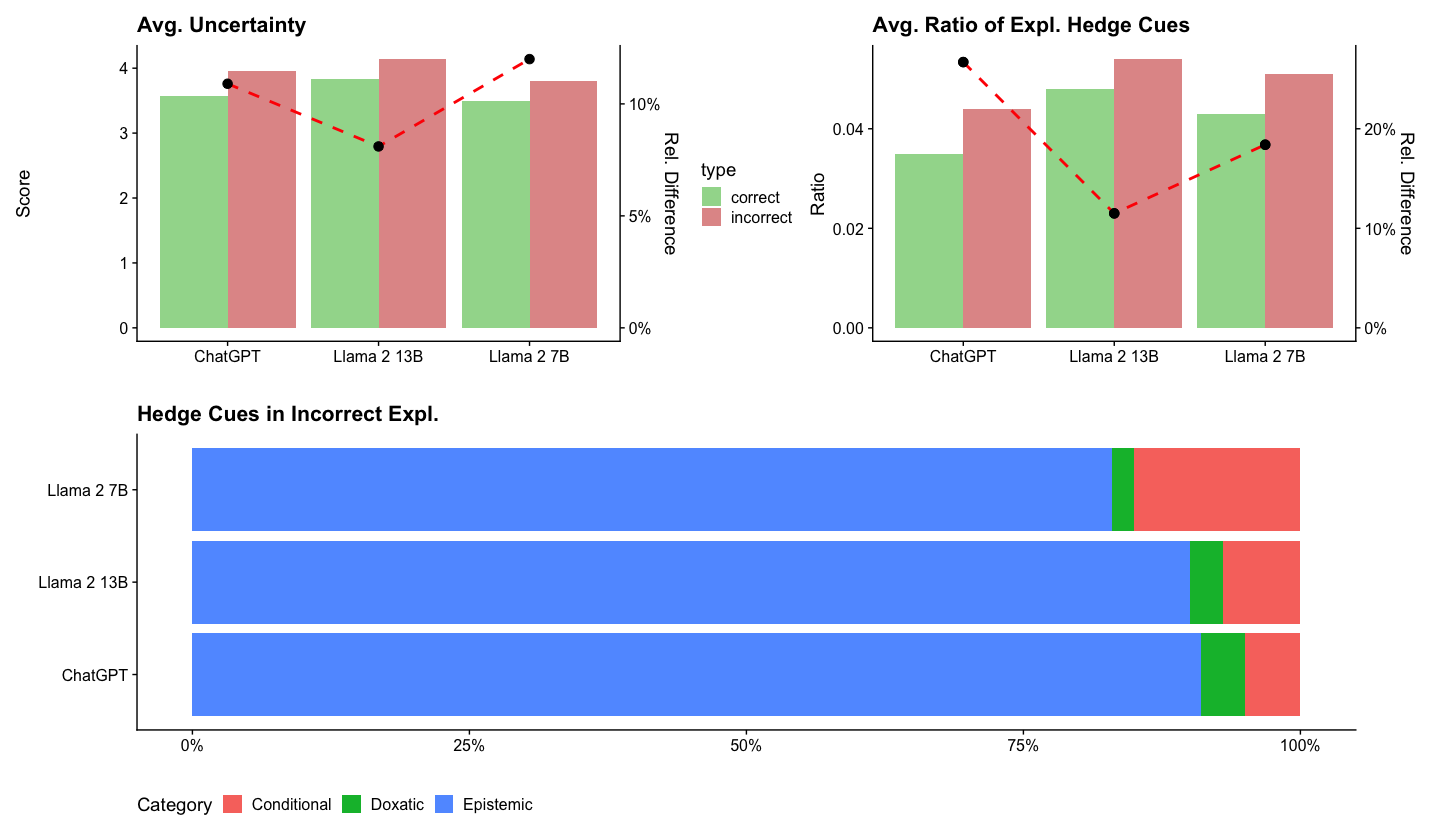}
    \caption{Evaluation of linguistic uncertainty in LLM-generated explanations. LLMs tend to use more hedging language in explanations supporting less plausible hypotheses. Across the LLMs, the hedging language is found to be predominantly \emph{epistemic} \ref{sec: appendix-uncertainty}.}
    \label{fig:hedge-ratio}
\end{figure}

The results reveal that \textbf{linguistic uncertainty is the strongest predictor of explanation quality and is a statistically significant feature for all LLMs}. This suggests that LLMs use more qualifying language when explaining weaker hypotheses (see Figure \ref{fig:hedge-ratio}). We found that \textbf{uncertainty can improve accuracy by 13pp on COPA and 4pp on E-CARE}. We also examine the uncertainty cues expressed by LLMs by analyzing both the frequency of hedge words and the types of hedge cues employed in incorrect explanations. We find the distribution of hedge cues across LLMs tends to be similar, with only minor differences between LLMs (Figure \ref{fig:hedge-ratio}). Epistemic cues were most frequently used by all three models, with LLaMA 2 7B being more likely to use conditional cues. See Appendix \ref{sec: appendix-uncertainty} for further details.\label{sec: appendix-uncertainty}.

\subsection{Correlation with Human Judgement. }
We first sample 100 generated explanation pairs across both the COPA and E-CARE tasks and evaluated LLMs. Two human evaluators are instructed to evaluate the pair of explanations and to select which explanation is most plausible. No additional information about the original question nor the correct answer is provided to prevent biasing the judge.  

The human evaluators on average were able to identify the explanation associated with the correct answer 96\% (COPA) and 91\% (E-Care) of time. We compute the inter-evaluator agreement score between two human evaluators and find that there is Cohen's Kappa score of .68 suggesting there is a strong agreement between the two evaluators. 

To evaluate if \textit{IBE-Eval} is correlated with human judgment, we compute the Spearman's rank correlation between GPT-3.5-as-a-judge, \textit{IBE-Eval} and human judgment. We find that GPT-3.5-as-a-judge exhibits a weak and statistically insignificant correlation with human judgment (0.31). \textbf{In contrast, we find that the \textit{IBE-Eval} is significantly aligned with human preferences (with a Spearman's correlation of 0.64 and p < 0.01) further suggesting the IBE's potential for automatic explanation evaluation. }

\section{Related Work}
Explorations of LLM reasoning capabilities across various domains (e.g. arithmetic, commonsense, planning, symbolic, etc) are an emerging area of interest \cite{llm-reason-survey1, llm-reason-survey2}. Prompt-based methods \cite{wei2022chain, zhou2023leasttomost, wang2023selfconsistency}, such as CoT, investigate strategies to elicit specific types of reasoning behavior through direct LLM interaction. \citet{olausson2023linc} investigate automatic proof generation and propose a neurosymbolic framework with an LLM semantic parser and external solver. \citet{creswell2022selectioninference} propose an inference framework where the LLM acts as both a selection and inference module to produce explanations consisting of causal reasoning steps in entailment tasks. Research on LLM faithfulness \cite{atanasova-etal-2023-faithfulness} investigates if LLM explanations are robust to spurious input alterations. \citet{parcalabescu2024measuring-self-consistency} propose a self-consistency measure CC-SHAP which measures how specific alterations to a model's input contribute to the generated explanation. This paper primarily draws inspiration from recent work on the evaluation of natural language explanations  \cite{quan2024enhancing,valentino-etal-2021-natural,wiegreffe2021teach,thayaparan2020survey,dalvi2021explaining,camburu2018snli}. However, differently from previous methods that require extensive human annotations or specific domain knowledge, we are the first to propose a set of criteria that can be automatically computed on explicit linguistic and logical features.

\section{Conclusion}
This paper proposed \textit{IBE-Eval}, an interpretable framework for LLM explanation evaluation inspired by philosophical accounts of Inference to the Best Explanation (IBE). \textit{IBE-Eval} can identify the best explanation supporting the correct answer with up to 77\% accuracy in CQA scenarios, improving upon a GPT 3.5 Judge baselines by +17\%. Our regression study suggests that LLM explanations tend to conform to IBE expectations and that \textit{IBE-Eval} is strongly correlated with human judgment. Linguistic uncertainty is the stronger IBE predictor for explanation quality closely followed by parsimony and coherence. However, we also found that LLMs tend to be strong conjecture models able to generate logically consistent explanations for less plausible hypotheses, suggesting limited applicability for the logical consistency criterion in isolation. We believe our findings can open new lines of research on external evaluation methods for LLMs as well as interpretability tools for understanding the LLM's underlying explanatory process. 


\section{Limitations}
\textit{IBE-Eval} offers an interpretable explanation evaluation framework utilizing logical and linguistic features. Our current instantiation of the framework is primarily limited in that it does not consider grounded knowledge for factuality. We observe that LLMs can generate factually incorrect but logically consistent explanations. In some cases, the coherence metric can identify those factual errors when the step-wise entailment score is comparatively lower. However, our reliance on aggregated metrics can hide weaker internal entailment especially when the explanation is longer or the entailment strength of the surrounding explanation steps is stronger. Future work can introduce metrics to evaluate grounded knowledge or perform more granular evaluations of explanations to better weight factual inaccuracies. 

Additionally, \textit{IBE-Eval} currently does not support single natural language explanations and was evaluated in the limited domain of causal commonsense reasoning. Future work will explore globally calibrating \textit{IBE-Eval} plausibility scores to extend evaluation to more diverse explanation generation and QA settings. Calibration efforts would allow for \textit{IBE-Eval} to generate comparable scores across unrelated explanations and could be used to produce global thresholds explanation classification.  

Finally, the list of criteria considered in this work is not exhaustive and can be extended in future work. However, additional criteria for IBE might not be straightforward to implement (e.g., unification power, hardness to variation) and would probably require further progress in both epistemological accounts and existing NLP technology.

\section{Ethics Statement}

The human annotators for computing the human judgment baseline are all authors of the papers and as such were not further compensated for the annotation task.

\section*{Acknowledgements}
This work was partially funded by the Swiss National Science Foundation (SNSF) project NeuMath (\href{https://data.snf.ch/grants/grant/204617}{200021\_204617}), by the EPSRC grant EP/T026995/1 entitled “EnnCore: End-to-End Conceptual Guarding of Neural Architectures” under Security for all in an AI-enabled society, by the CRUK National Biomarker Centre, and supported by the Manchester Experimental Cancer Medicine Centre,
the Science Foundation Ireland under grants SFI/18/CRT/6223 (Centre for Research Training in Artificial Intelligence), SFI/12/RC/2289\_P2 (Insight), co-funded by the European Regional Development Fund,
and the NIHR Manchester Biomedical Research Centre.
\bibliography{custom}
\bibliographystyle{acl_natbib}

\appendix

\section{Appendix}

\label{sec:appendix}

\subsection{Reproducibility}
All experimental code is available online\footnote{\url{https://github.com/dhairyadalal/IBE-eval}} to encourage future research in the field. We additionally summarize all the model implementations and technical resources used for the computation of the proposed IBE criteria below:
\begin{itemize}
  \item We adopt the Prolog solver for neuro-symbolic integration from \cite{quan2024enhancing}.
  \item  We use Spacy \cite{spacy2} to tokenize and extract part-of-speech (POS) tags.
  \item To compute coherence, we employ the RoBERTa-based NLI model \cite{nie-etal-2020-adversarial} that has been finetuned on a range of NLI and fact verification datasets consisting of SNLI \cite{snli}, aNLI \cite{nie-etal-2020-adversarial}, multilingual NLI \cite{mnli}), and FEVER-NLI \cite{fever}.
  \item To measure sentence-level uncertainty, we employ a finetuned RoBERTa model provided by \cite{pei-jurgens-2021-measuring}.
  \item We use a fine-tuned BERT-based token classification model to classify all the words in the generated explanation with uncertainty categories introduced in the 2010 CoNLL shared task on Hedge Detection \cite{farkas-etal-2010-conll}. 
\end{itemize}

\subsection{Explanation Prompting}
\label{sec:appendix-prompting}
\begin{figure}[h!]
    \centering
    \includegraphics[width=.5\textwidth]{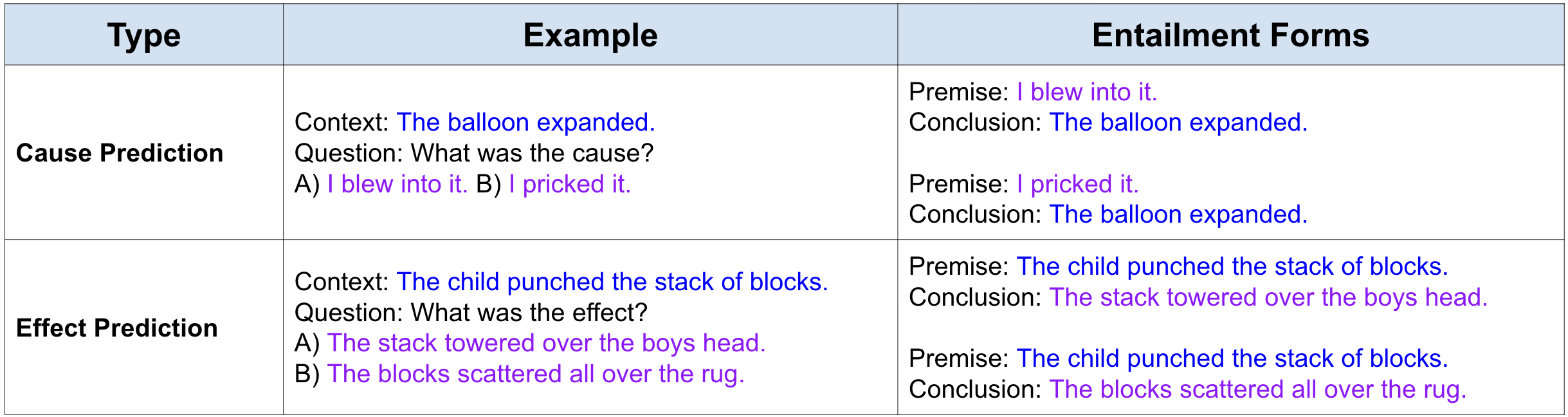}
    \caption{To perform IBE we convert the CQA context and answer candidates into an entailment form (i.e., EEV) \cite{valentino-etal-2021-natural}. }
    \label{fig:eev}
\end{figure}

\begin{figure}[!]
    \centering
    \includegraphics[width=.5\textwidth]{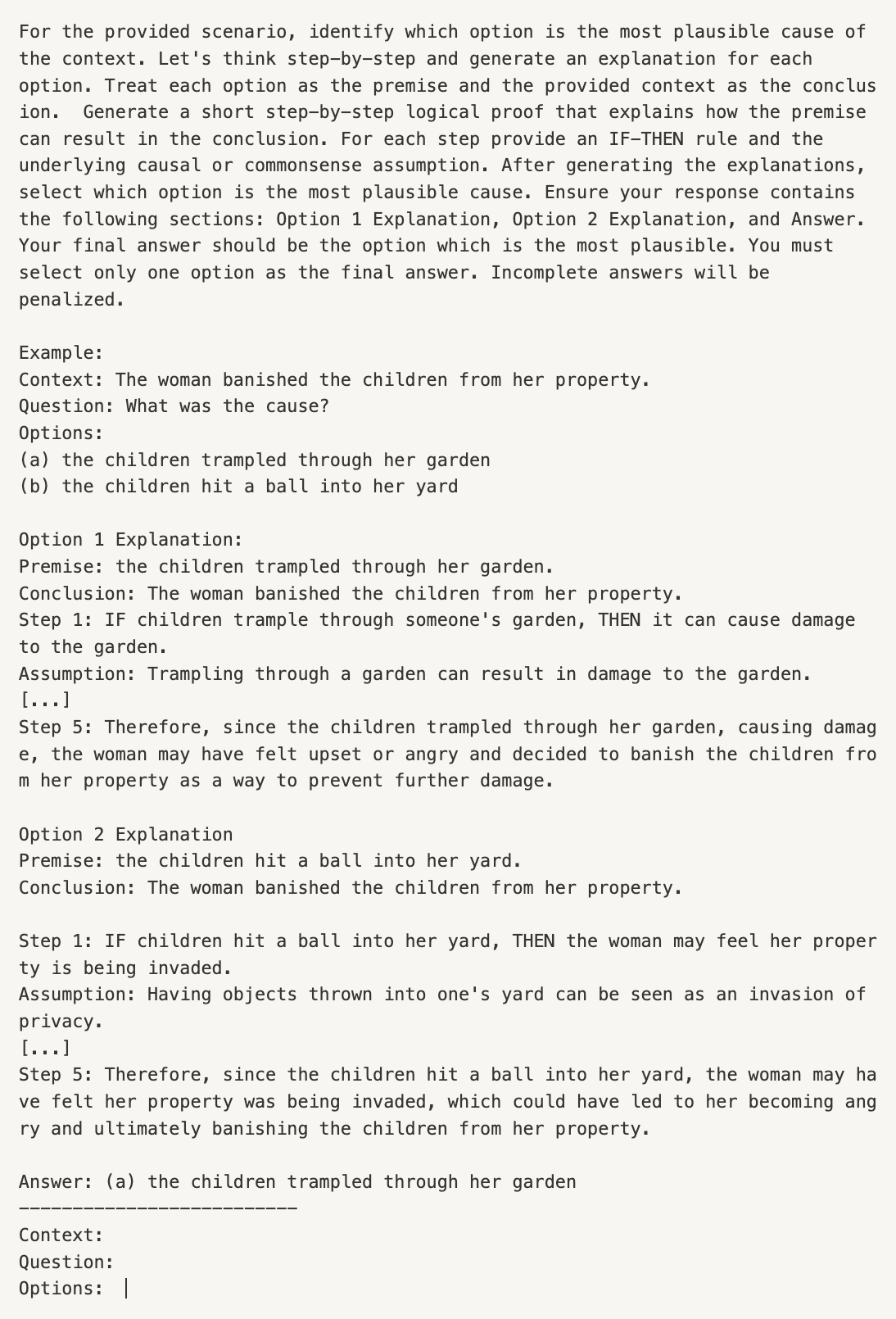}
    \caption{An example of the modified CoT prompt template for explanation generation.}
    \label{fig:llm-prompting}
\end{figure}

A modified CoT prompt is used to instruct the LLM to generate explanations. The prompt includes a set of instructions for explanation generation and an in-context example. Appended to the end of the prompt are the CQA context, causal question, and answer candidates. The LLM is instructed to first convert the options into the EEV format consisting of a premise and conclusion. The EEV format will differ depending on the directionality of the causal question (see Figure \ref{fig:eev}). Cause prediction questions will treat the answer candidate as the premise and the context as the conclusion. In contrast, effect prediction reverses the relationship treating the context as the premise and the answer options as the conclusion. After the EEV conversion, the model is instructed to generate a step-by-step explanation consisting of IF-THEN statements and the associated causal or commonsense assumptions. For ease of post-processing, the LLM is instructed to use headers and enumerate steps using the \textit{Step \#} format. A full example of the prompt template is exhibited in Figure \ref{fig:llm-prompting}.

\begin{figure}[!]
    \centering
    \includegraphics[width=.5\textwidth]{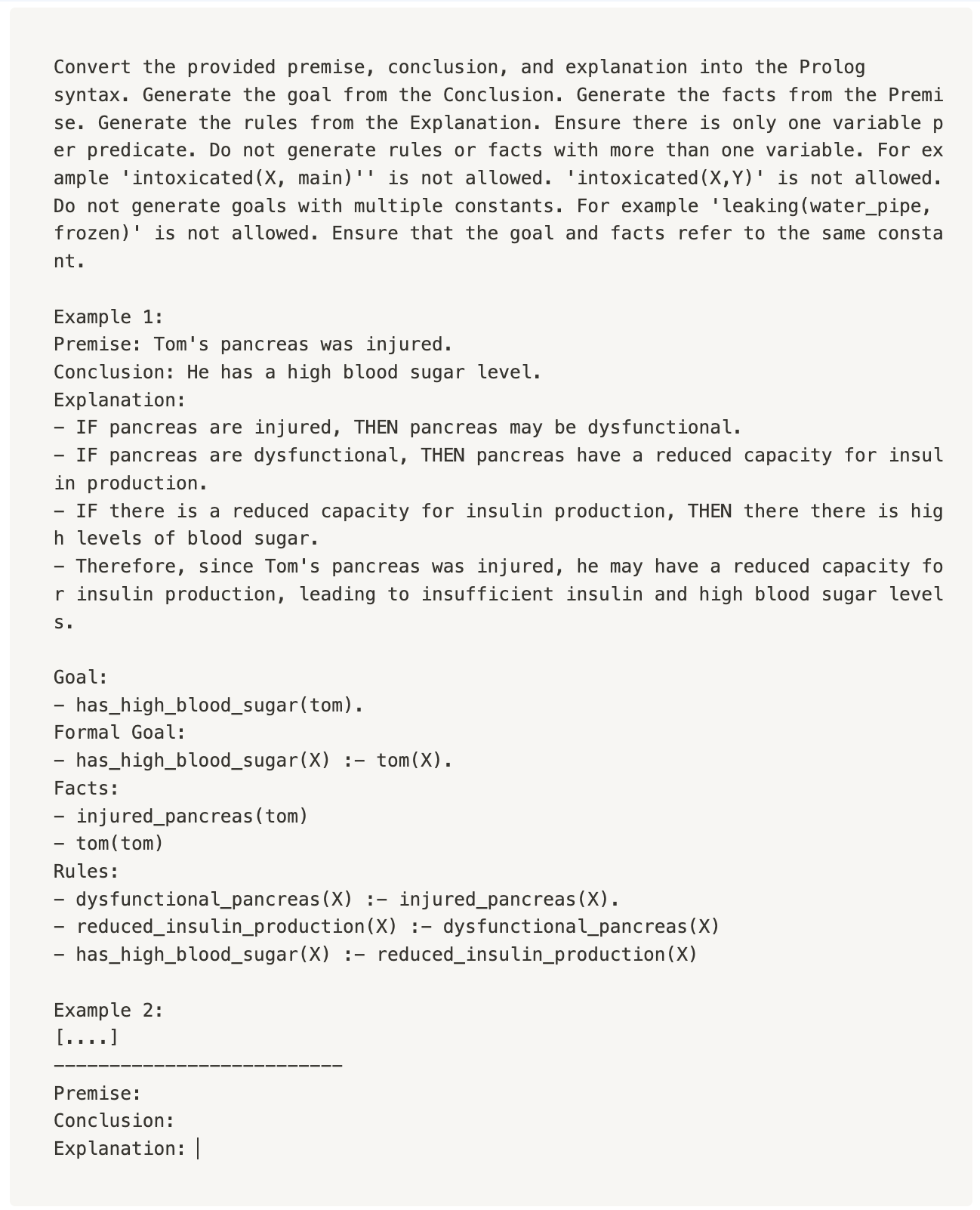}
    \caption{An example of the autoformalization prompt.}
    \label{fig:autoformalization}
\end{figure}

\subsection{Autoformalization}
\label{sec:appendix-autoformalization}
Autoformalization is the process of translating natural language descriptions into formal specifications \cite{wu2022autoformalization}. We adopt the translational capabilities of GPT-3.5-Turbo to convert the explanation into a formal entailment hypothesis. The IF-THEN explanation steps are converted into a set of Prolog rules, the entailment description is used to generate Prolog atoms, and the conclusion statement is translated into a Prolog query. We provide an example of the autoformalization prompt in Figure \ref{fig:autoformalization} and an example of the formalized output in Figure \ref{fig:logical-consistency}. After autoformalization, we deploy a post-processing script to extract the formalized rules, atoms, and query and generate a Prolog program for entailment verification.


\subsection{LLM-as-a-Judge Baseline}
\label{sec:appendix-llm-judge}
\begin{figure}[h!]
    \centering
    \includegraphics[width=.5\textwidth]{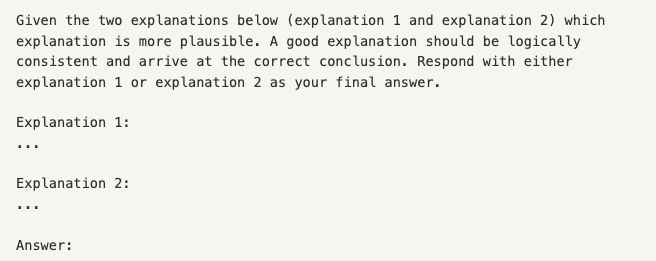}
    \caption{An example of prompt used by the LLM-as-a-Judge model for evaluating competing explanations.}
    \label{fig:logical-consistency}
\end{figure}

GPT 3.5 is used as the LLM for the LLM-as-a-Judge baseline. Similar to the human evaluators, GPT is presented with both generated explanations and asked to identify which explanation is more plausible.

\subsection{Logical Consistency} 
\label{sec:appendix-consistency}
\begin{figure}[h!]
    \centering
    \includegraphics[width=.5\textwidth]{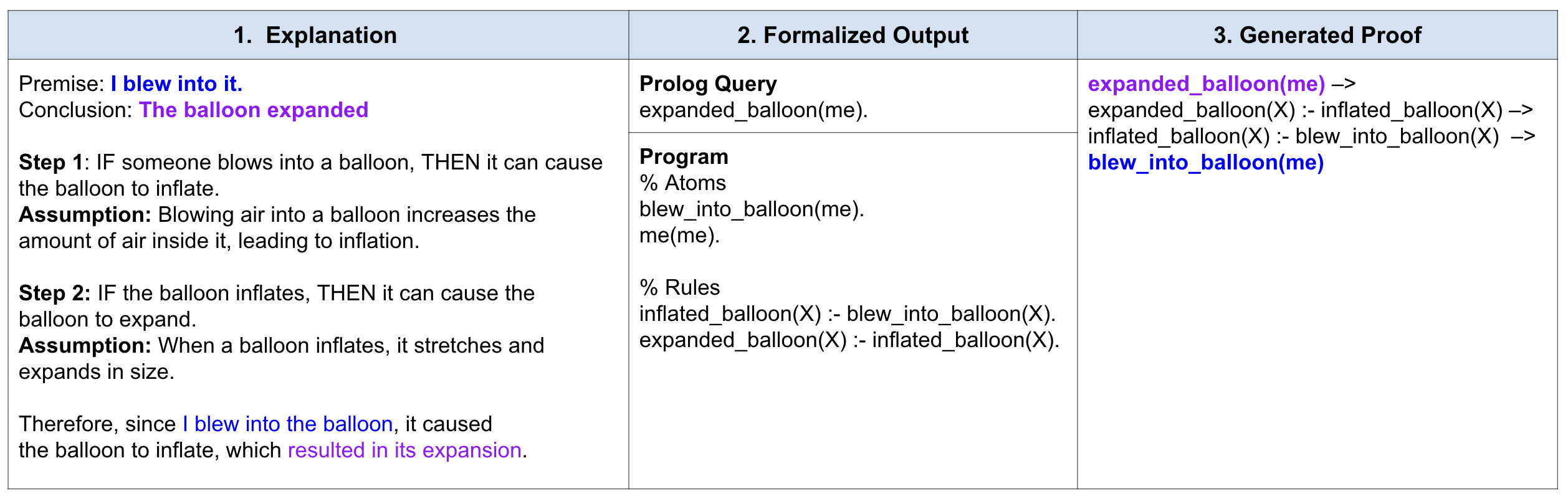}
    \caption{An example of the autoformalization prompt.}
    \label{fig:logical-consistency}
\end{figure}
An explanation hypothesis is considered logically consistent if the external solver can build a deductive proof connecting the conclusion to the premise. We use NLProlog \cite{weber2019nlprolog}, a neuro-symbolic Prolog solver integrating backward chaining with word embedding models via a weak unification mechanism. NLProlog allows for a level of flexibility and robustness that is necessary for NLP use cases (e.g. unification applied to synonyms). We provide the autoformalized query, atoms, and rules to NLProlog. If NLProlog can satisfy the entailment query, it will return the proof consisting of the set of rules traversed, the weak unification score, and the proof depth. For simplicity, we assign a score of one if the entailment query is satisfied and zero if it is not. The proof depth score is evaluated as part of the parsimony analysis. An end-to-end example of consistency evaluation can be found in Figure \ref{fig:logical-consistency}.

\begin{algorithm}[h!]
    \SetKwFunction{isOddNumber}{isOddNumber}
    \SetKwInOut{KwIn}{Input}
    \SetKwInOut{KwOut}{Output}
    
    \KwIn{Symbolic KB $kb$, Goal $goal$, Glove embedding model $e(\cdot)$}
    \KwOut{proof chain $chain$, proof depth $depth$}
    
    \BlankLine
    threshold $\leftarrow$ 0.13\;
    
    \BlankLine
    $depth$ $\leftarrow$ 1 \;
    $chain$ $\leftarrow$ $empty list$ \;
    
    \ForEach{step $t$ {\bf in} backward\_chaining($kb$, $goal$)}{
        \ForEach{$max\_unification(q, q_t)$}{
            $unification\_score$ $\leftarrow$ $CosineSimilarity$($e(q, m_s), e(q_t, m_s)$)\;
            $depth$ $\leftarrow$ $depth$ $\times$ $unification\_score$ \;
        }
          $chain$ $\leftarrow$ backward\_chaining($kb$, $goal$)\;
    }
    
    \BlankLine
    
    \If{$chain$ is not empty {\bf and} $depth$ $>$ threshold}{
        $chain \leftarrow$ current\_proof\_chain$[0]$\;
    }
    \Else{$depth$ $\leftarrow$ 0 \;}{}
    
    \BlankLine
    \KwRet{$chain$, $depth$ }\;
    \caption{Neuro-symbolic Solver}
    \label{alg: appendix-solver}
\end{algorithm}

\subsection{Parsimony}
\label{sec: appendix-parsimony}
Parsimony measures the complexity of an explanation and is represented by the proof depth and concept drift metrics. Proof depth is automatically calculated by NLProlog and reflects the number of rules traversed by the solver to satisfy the entailment query. If the hypothesis is not logically consistent, depth is set to zero. The concept drift metric measures the entropy of novel concepts introduced to bridge the premise and conclusion. To compute the drift of an explanation, we consider the nouns found in the premise, conclusion, and explanation steps. We use Spacy \cite{spacy2} to tokenize and extract part-of-speech (POS) tags. All tokens with the 'NOUN' POS tag extracted. For normalization purposes, we consider the lemma of the tokens. Concept drift then is calculated as the set difference between the unique nouns found across all explanation steps and those found in the premise and conclusion.

\begin{algorithm}[h!]
    \SetAlgoLined
    \caption{Concept Drift}

    \SetKwInOut{KwIn}{Input}
    \SetKwInOut{KwOut}{Output}
    
    \KwIn{Premise, Conclusion, Explanation, Spacy model $spacy$($\cdot$)}
    \KwOut{Drift Score $drift$}
    
    \BlankLine
    $Noun_{p} \leftarrow$ $spacy$($Premise$)\;
    $Noun_{c} \leftarrow$ $spacy$($Conclusion$)\;
    $Noun_{E} \leftarrow$ $spacy$($Explanation$)\;
    $N \leftarrow \{Noun_{p}, Noun_{c}, Noun_{E}\}$\;
    $drift \leftarrow length(set(Noun_{E}) - set(Noun_{p} \cup Noun_{c})$)\;
    \Return{$drift$}\;
    \label{alg: appendix-drift}
\end{algorithm}

\subsection{Coherence}
\label{sec: appendix-coherence }
Coherence evaluates the plausibility of the intermediate explanation. We propose stepwise entailment as a metric to measure the entailment strength of the If-then implications. We employ a RoBERTa-based NLI model \cite{nie-etal-2020-adversarial} that has been finetuned on a range of NLI and fact verification datasets consisting of SNLI \cite{snli}, aNLI \cite{nie-etal-2020-adversarial}, multilingual NLI \cite{mnli}), and FEVER-NLI \cite{fever}. To compute the stepwise entailment score, we first measure the entailment strength between the If and Then propositions. For example, to calculate the score of the statement \emph{``IF a balloon is pricked, THEN the balloon may deflate''} we consider \emph{``a balloon is pricked''} and \emph{``the balloon may deflate''} as input sentences for the NLI model. The NLI will produce independent scores for the entailment and contradiction labels. We compute the entailment strength by subtracting the contraction label score from the entailment label score. An entailment strength of one indicates the If-then implication is strongly plausible whereas a score of zero suggests that it is likely implausible. The overall stepwise entailment score is the average of entailment strength measures across all explanation steps. 

\begin{algorithm}[h!]
    \SetAlgoLined
    \caption{Stepwise Entailment}
    
    \SetKwInOut{KwIn}{Input}
    \SetKwInOut{KwOut}{Output}
    
    \KwIn{Explanation $E$, NLI Model $nli$($\cdot$)}
    \KwOut{Average Entailment Strength $strength$}
    
    \BlankLine
    $EntailmentStrengthScores \leftarrow$ empty list\;
    
    \BlankLine
    \ForEach{Step $(If_s, Then_s)$ in $E$}{
        \BlankLine
        $EntailmentScore \leftarrow$ $nli$($If_s$, $Then_s$)\;
        $ContradictionScore \leftarrow$ $nli$($If_s$, $Then_s$)\;
        $EntailmentStrength \leftarrow EntailmentScore - ContradictionScore$\;
        Append $EntailmentStrength$ to $EntailmentStrengthScores$\;
    }
    
    \BlankLine
    $strength \leftarrow$ $Avg$($EntailmentStrengthScores$)\;  
    \BlankLine
    \Return{$strength$}\;
\end{algorithm}

\subsection{Linguistic Uncertainty}
\label{sec: appendix-uncertainty}

Linguistic uncertainty measures the confidence of a statement where hedging cues and indirect language suggest ambiguity around the proposition. To measure sentence-level uncertainty, we employ a finetuned RoBERTa model provided by \cite{pei-jurgens-2021-measuring}. The model was trained on a sentence-level dataset consisting of findings and statements extracted from new articles and scientific publications and human annotated evaluation of sentence certainty. A scale from one to six was used to annotate sentences where one corresponds to the lowest degree of certainty and six is the highest expressed by the sentence. We invert the scale to retrieve the uncertainty scores. To compute the overall linguistic uncertainty of an explanation, we first compute the uncertainty for each assumption and the explanation summary and then average all the scores. 

 We use a fine-tuned BERT-based token classification model to classify all the words in the generated explanation with uncertainty categories introduced in the 2010 CoNLL shared task on Hedge Detection \cite{farkas-etal-2010-conll}. \citet{farkas-etal-2010-conll} classify hedge cues into three categories: epistemic, doxatic, and conditional. Epistemic cues refer to hedging scenarios where the truth value of a proposition can be determined but is unknown in the present (e.g. the blocks \textit{may} fall). Doxatic cues refer to beliefs and hypotheses that can be held to be true or false by others (e.g. the child \textit{believed } the blocks \textit{would} fall). Finally, conditional cues refer to propositions whose truth value is dependent on another proposition's truth value (e.g. \textit{if} the balloon is pricked it may deflate).
 
\begin{algorithm}[h!]
    \SetAlgoLined
    \caption{Linguistic Uncertainty}
   
    \SetKwInOut{KwIn}{Input}
    \SetKwInOut{KwOut}{Output}
    
    \KwIn{Assumptions, Explanation Summary, Uncertainty Estimator Model $uc(\cdot)$}
    \KwOut{Overall Uncertainty}

    \BlankLine
    $AssumptionUncertaintyList \leftarrow$ empty list\;
    
    \BlankLine
    \ForEach{Assumption in Assumptions}{
        \BlankLine
        $UncertaintyScore \leftarrow$ $uc$($UncertaintyModel$, $Assumption$)\;
        Append $UncertaintyScore$ to $AssumptionUncertaintyList$\;
    }
    
    \BlankLine
    $AverageAssumptionUncertainty \leftarrow$ 
    $Avg$($AssumptionUncertaintyList$)\;
    
    \BlankLine
    $ExplanationUncertainty \leftarrow$ $uc$($UncertaintyModel$, $ExplanationSummary$)\;
    
    \BlankLine
    $OverallExplanationUncertainty \leftarrow AverageAssumptionUncertainty + ExplanationUncertainty$\;
    
    \BlankLine
    \Return{$OverallExplanationUncertainty$}\;
\end{algorithm}

\subsection{Inference to Best Explanation}
\label{sec: appendix-ibe}

To perform IBE, we first fit a linear regression model over the extracted explanation features from the COPA train set and 500 random sample train examples from the E-CARE train set. We consider all explanations independently and annotate each explanation with a 1 if it corresponds to a correct answer or 0 if corresponds to an incorrect answer.  After the linear model is fitted, we evaluate the COPA and E-CARE test sets. For each example, we use the trained linear model to score each answer candidate explanation and then select a candidate with the highest score. We use the linear regression implementation from scikit-learn \cite{sklearn_api} for the IBE model. We additionally use the R stats package \cite{rstats} for conducting our regression analysis. 

\subsection{E-CARE Results}

\label{sec: appendix-ecare-results}
\subsubsection{E-CARE Consistency}
\begin{figure}[t!]
    \centering
    \includegraphics[width=.4\textwidth]{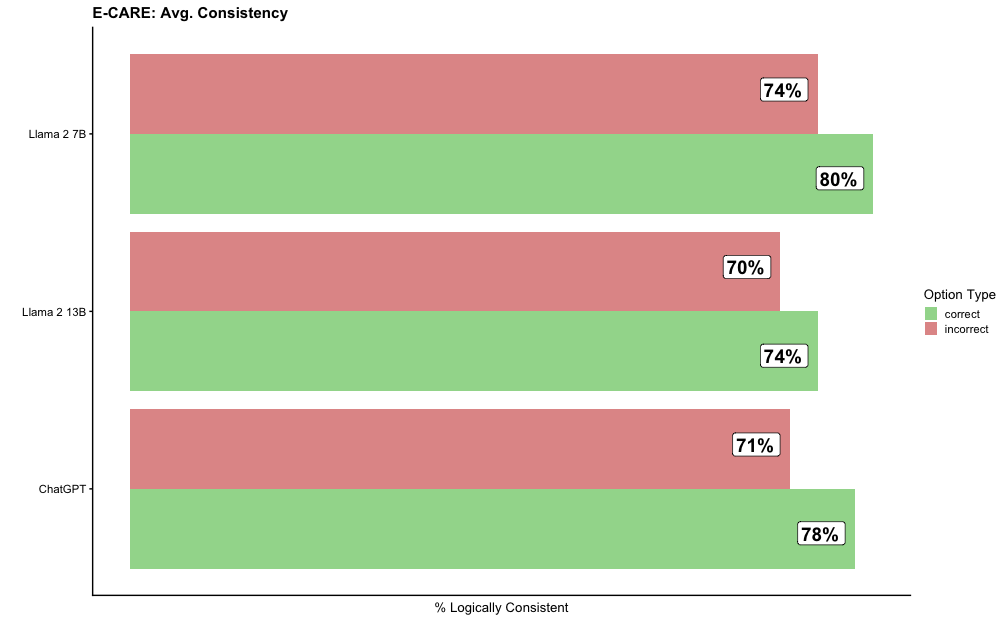}
    \caption{Average consistency comparison between correct and incorrect options for the E-CARE dataset.}
    \label{fig:ecare-consistency}
\end{figure}

See Figure \ref{fig:ecare-consistency}.

\subsubsection{ E-CARE Proof Depth}
\begin{figure}[t!]
    \centering
    \includegraphics[width=.4\textwidth]{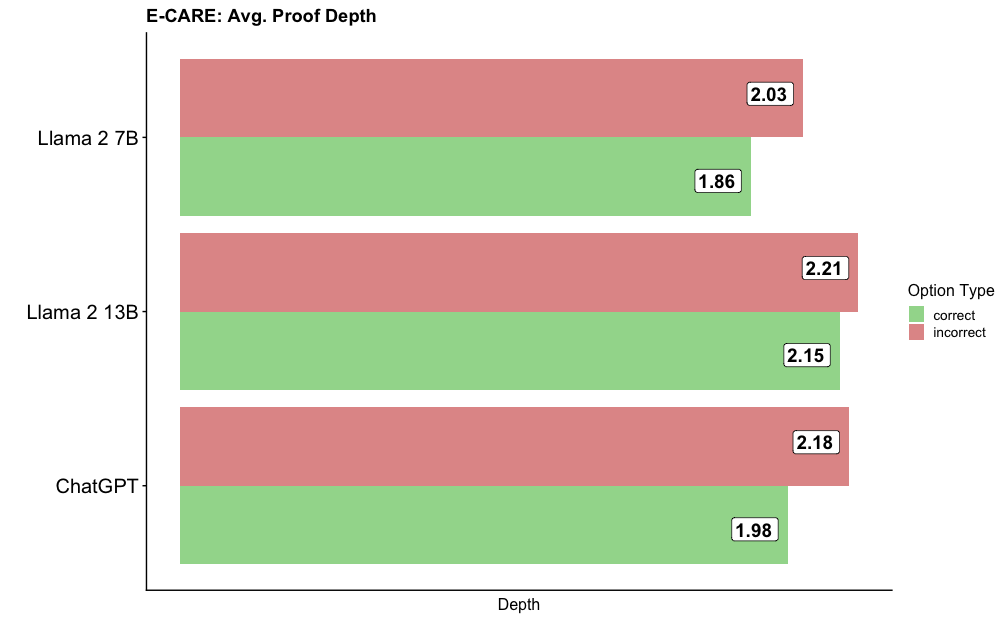}
    \caption{Comparison of average proof depth between correct and incorrect options.}
    \label{fig:ecare-depth}
\end{figure}

See Figure \ref{fig:ecare-depth}.

\begin{figure}[t!]
    \centering
    \includegraphics[width=.4\textwidth]{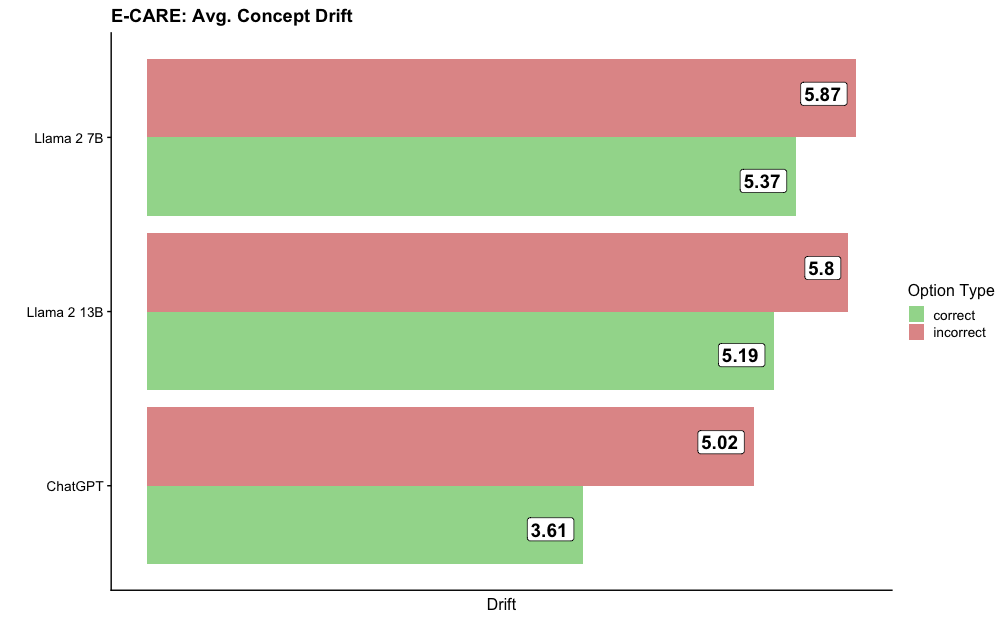}
    \caption{Comparison of average concept drift between correct and incorrect options.}
    \label{fig:ecare-drift}
\end{figure}
\begin{figure}[t!]
    \centering
    \includegraphics[width=.4\textwidth]{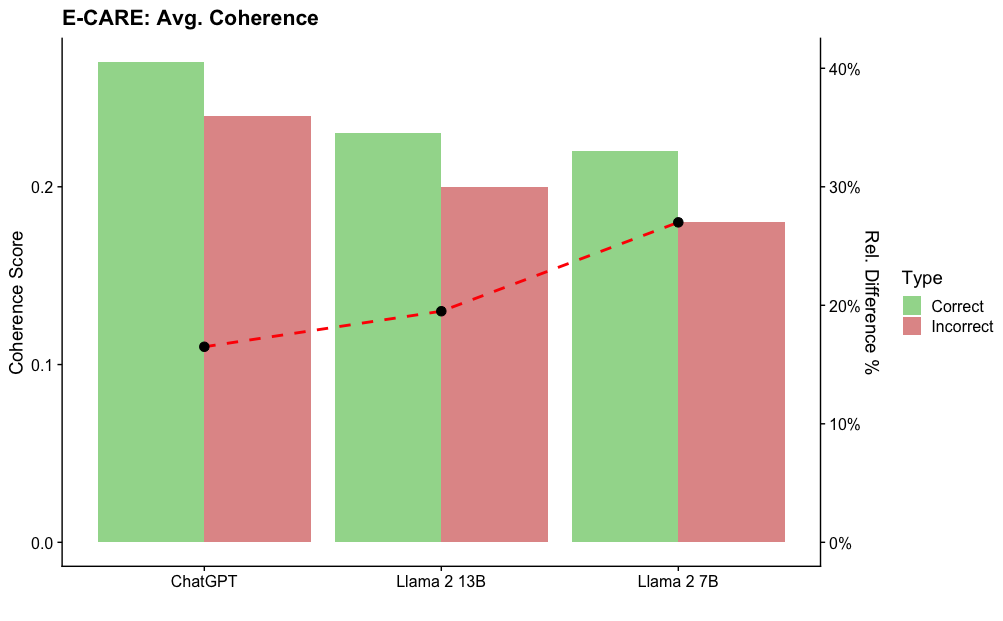}
    \caption{Comparison of average coherence scores between correct and incorrect options.}
    \label{fig:ecare-coherence}
\end{figure}

\begin{figure}[h!]
    \centering
    \includegraphics[width=.4\textwidth]{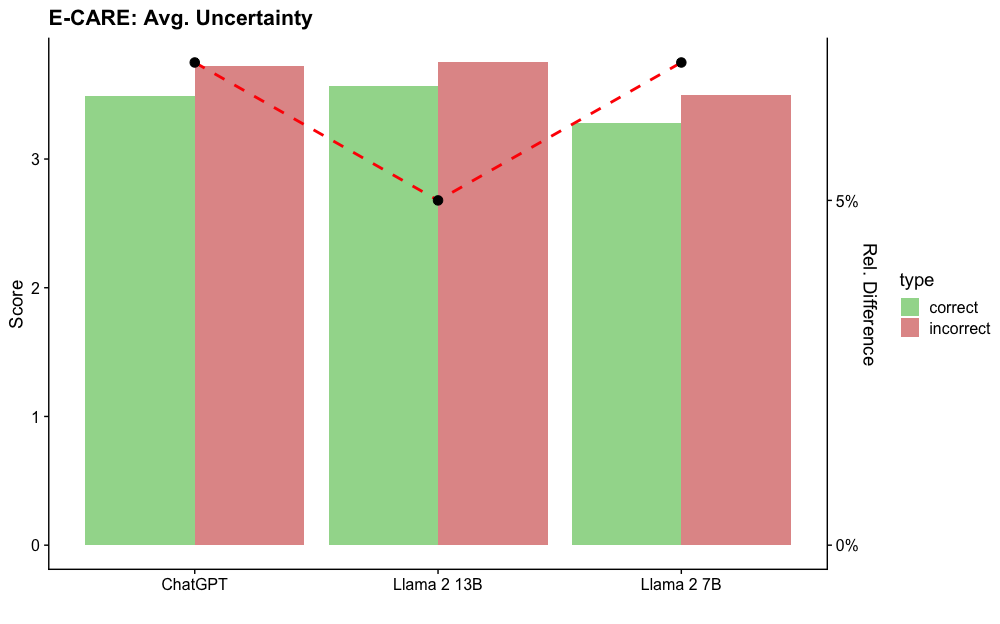}
    \caption{Comparison of average uncertainty scores between correct and incorrect options.}
    \label{fig:ecare-uncertainty}
\end{figure}

\begin{figure}[h!]
    \centering
    \includegraphics[width=.4\textwidth]{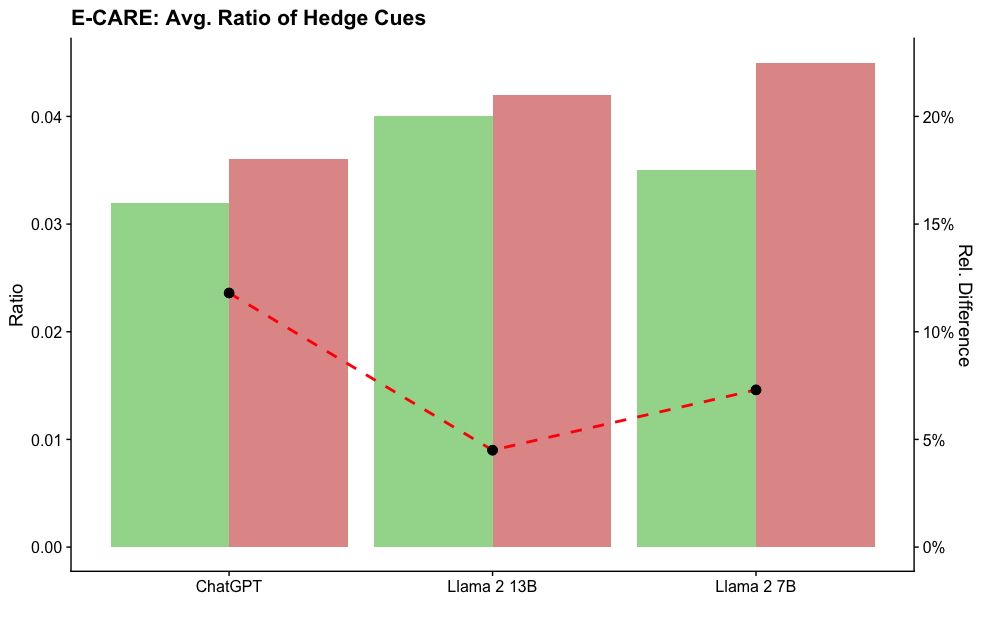}
    \caption{Comparison of the average ratio of hedge cues between correct and incorrect options.}
    \label{fig:ecare-hedge-ratio}
\end{figure}

\begin{figure}[h!]
    \centering
    \includegraphics[width=.4\textwidth]{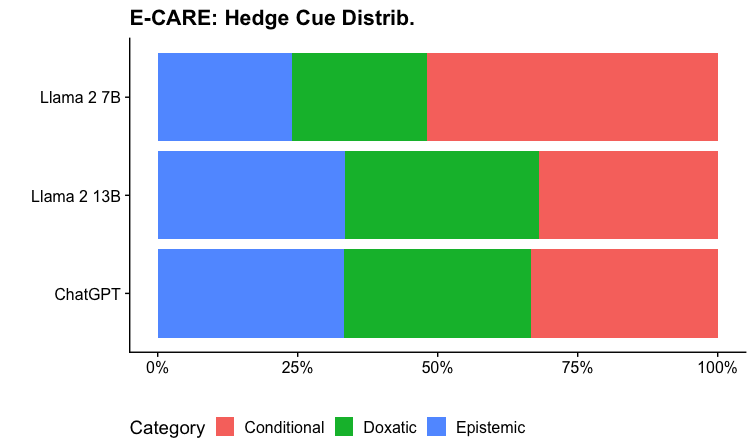}
    \caption{Distribution of hedge cues across incorrect explanations.}
    \label{fig:ecare-hedge-distribution}
\end{figure}

\subsubsection{ E-CARE Concept Drift}
See Figure \ref{fig:ecare-depth}.
\subsubsection{ E-CARE Coherence}
See Figure \ref{fig:ecare-coherence}.
\subsubsection{ E-CARE Uncertainty}
See Figure \ref{fig:ecare-uncertainty}.
\subsubsection{ E-CARE Hedge Ratio}
See Figure \ref{fig:ecare-hedge-ratio}.
\subsubsection{ E-CARE Hedge Distribution}
See Figure \ref{fig:ecare-hedge-distribution}.

\subsection{Causal Directionality}
\begin{figure}[t]
    \centering
    \includegraphics[width=.4\textwidth]{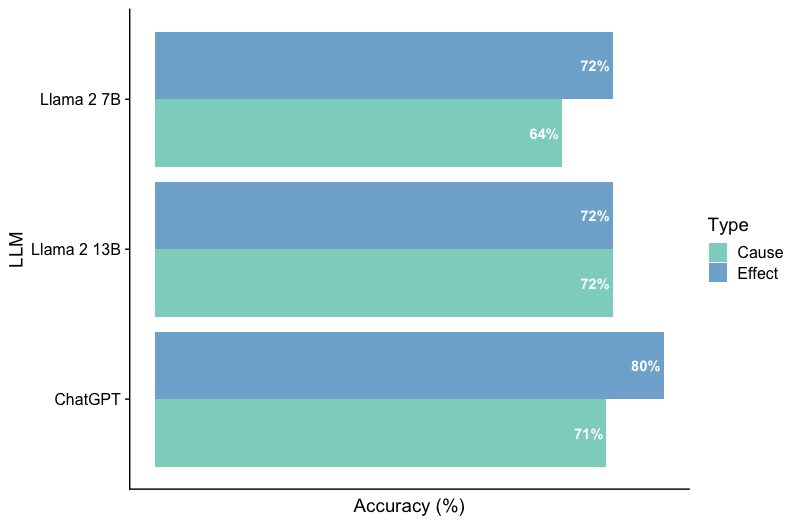}
    \caption{Accuracy in predicting the most plausible causes vs effects on COPA.}
    \label{fig:cause-effect}
\end{figure}

When considering the causal directionality (i.e. cause vs effect), we observed that accuracy tended to differ between LLMs on COPA. In particular, \textbf{we found both GPT and LLaMA 2 7B to be more accurate in predicting the effects in causal scenarios} (see Figure \ref{fig:cause-effect}). We hypothesize that \textbf{LLMs may suffer the challenge of causal sufficiency as the space of potential causal explanations can be far greater than the range of effects once an event has been observed}. This hypothesis is partly supported by the fact that GPT and LLaMA 2 7B express greater linguistic uncertainty and produce more complex explanations when predicting causes rather than effects. 

\subsection{Dataset Details}
COPA is released under a BSD-2 license and made available for broad research usage with copyright notification restrictions \footnote{\url{people.ict.usc.edu/~gordon/copa.html}}. We do not modify or use COPA outside of its intended use which is primarily open-domain commonsense causal reasoning. E-CARE is released under the MIT license and can be used for broad purposes with copyright notification restrictions \footnote{\url{github.com/Waste-Wood/e-CARE?tab=MIT-1-ov-file\#readme}}. We do not modify or use E-CARE outside of its intended use which is causal reasoning evaluation of language models.

\end{document}